# PetLock：A Genderless and Standard Interface for the Future On-orbit Construction

*Yuntao Li[1], Zichun Xu[1], Xiaohang Yang[1], Zhiyuan Zhao[1], Jingdong Zhao\*,[1], Hong Liu[1]*

1  *State Key Laboratory of Robotics and System, Harbin Institute of Technology, Harbin, Heilong Jiang, China*

**Abstract**: Modular design is the foundation of on orbit construction technology of large space facilities in the future. Standard interface is the key technology of modular design of the future space robotic systems and space facilities. This paper presents the designed and tested of **PetLock**, a standard and genderless interface which can transfer mechanical loads, power and data between the future modular space robotic manipulator and spacecraft. **PetLock** adopts a completely genderless design, including connection face, locking mechanism, data and power interface. The connection surface provides a large translation and rotation misalignment tolerance, due to its 120-degree symmetrical and 3D shape design. The locking mechanism features the three locking pins retraction structure design, which is simple and reliable. POGO pin connectors in the center of the interface provides the power and data transfer capabilities. Due to the advantages of high locking force, large tolerance, high reliability and low cost, **PetLock** has the very big application potential in future on orbit construction missions.

## 1. Introduction

With the development of space science and technology, the construction of large-scale space facilities such as space telescopes, large-scale communication antennas and space solar stations is increasingly urgent. However, due to the constraints of rocket carrying capacity, fairing envelope and structural complexity, large space facilities cannot be launched into orbit at one time. On orbit assembly technology can break through the limitations of vehicles and make it possible to build super large space facilities. On orbit assembly refers to the assembly of large space facility modules (LSFM，Fig. 1) into large space facilities through standard interfaces in space.

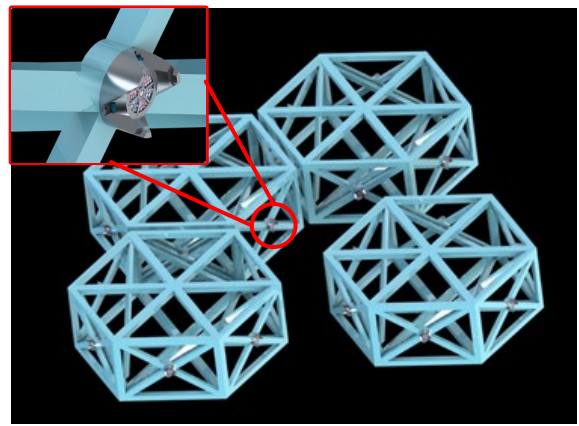

Fig.1 The assembly of large space facility modules

On orbit construction (OOC) experience of the international space station (ISS) shows that space robots are the core equipment of on orbit construction, which can greatly improve the safety and economy of space machine operation[1]. However, the current space robot system can not meet the needs of future OOC missions. Compared with the ISS, the OOC of large-scale space facilities in the future will require more workload, more types of tasks and more complex working environment. The modular reconfigurable space robotic systems (MRSRS), which can be configured to the desired configuration and number of branches through joints and telescopic arms with standard interfaces according to the environment and task requirements (Fig. 2 (a)). The MRSRS also can replace the required end effectors (Fig. 2 (b)), crawl on the space truss (Fig. 2 (c)), directly operate the LSFM by the standardized interfaces (Fig. 2 (d)). The MRSRS have great application prospects in the future on orbit construction.

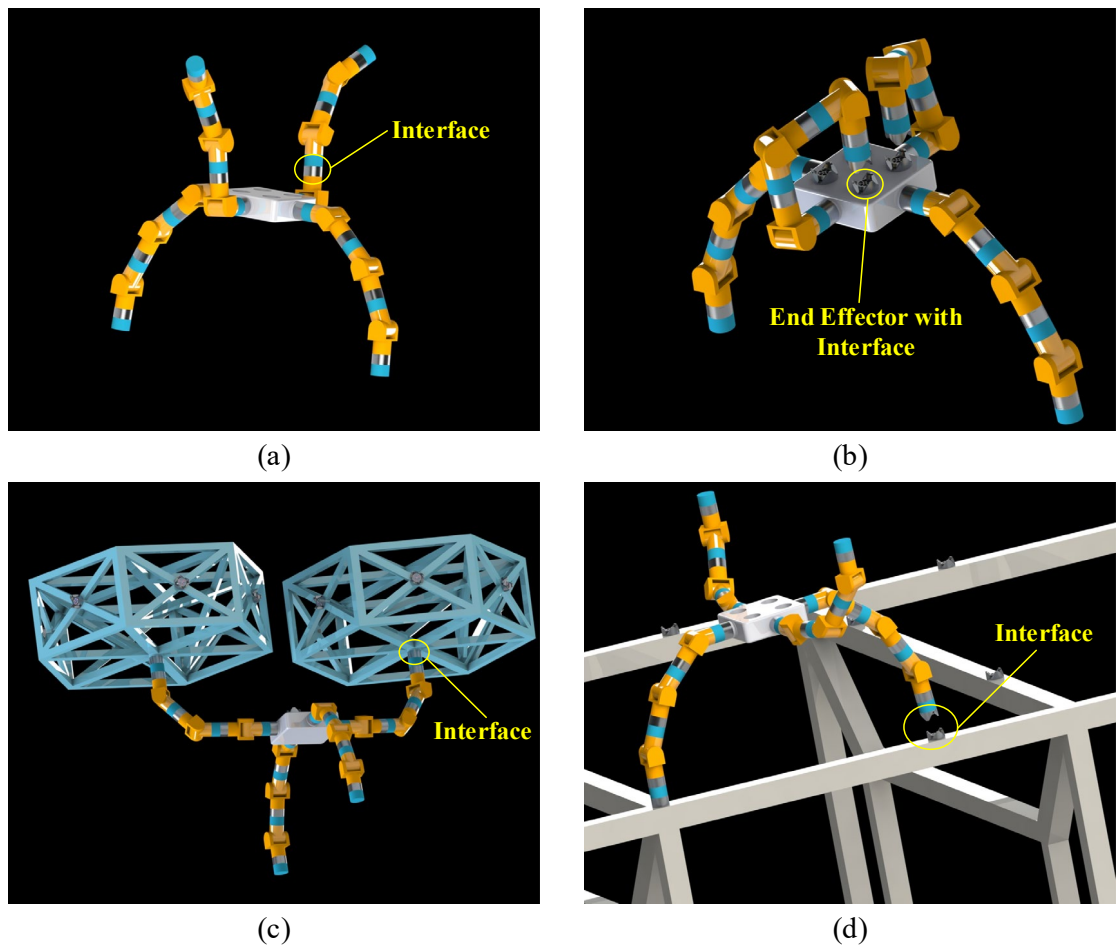

Fig.2 The application of standard interface

Standardized interface is the core technology of modular design of space facilities and space robots. Europe has conducted in-depth research on standardized interfaces, among which the representative ones are ISSI[2] 、SIROM[3] and HOTDOCK[4]. Table 1 compares the characteristics and main features of the three interfaces.

This paper develops ***PetLock*** (Fig.3), a genderless and standard interface with large misalignment tolerance, high connection strength, high reliability and light weight, which can transfer mechanical loads, power and data. Installed on the LSFM, the LSFM can be assembled into a large space facility. Installed on the robotic manipulator, the robotic manipulator can change the configuration, grab the space facility module, and replace the end effectors. Installed on the space truss, the space robot can crawl on the space truss. ***PetLock*** shows great application potential.

Table 1 The comparison of main features of three interfaces

|  | ISSI | SIROM | HOTDOCK |
|---|---|---|---|
| **Dimensions (diameter x height)** | Active: 143.7 x 54.2 mm | 128 x 76.6 mm | Active: 148 x 70 mm<br>Passive: 120 x 35 mm<br>Mechanical: 120 x 25mm |
| **Mass** | Active: 1kg<br>Passive: 0.65/0.2kg | Single:1.5kg | Active: 1.55kg<br>Passive: 0.5 kg<br>Mechanical: 0.25kg |
| **General features** | Gender or Genderless, 90° symmetry | Genderless,120° symmetry | Gender, 90° symmetry |
| **Misalignment tolerances** | -- | +/- 10 mm in axial direction<br>+/- 5mm in other axes<br>1.5° in rotation | +/- 15mm in translation<br>10° in rotation |
| **Load transfer (nondestructive)** | 6,000N axial, 400N lateral, 100Nm bending/torque | 1020 N in traction | 3000 N in traction<br>300 Nm in bending moment |
| **Coupling Time** | 5-12S | 60S | 20s |
| **Power transfer** | 5kW@100V | 120 W for 100 V line<br>30 W for 24 V line | 2.5 kW @ 120V (up to 4kW with specific pin configuration) |
| **Data transfer** | CAN, Ethernet | CAN, SpW | CAN, SpW, Ethernet, TTE |

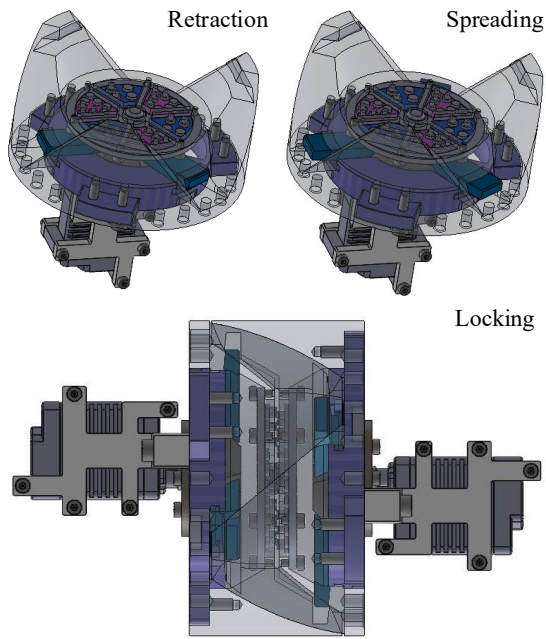

Fig.3 The CAD model of *PetLock*

The following structure of this paper is as follows: Section 2 introduces the specific design of *PetLock*. Section 3 verifies the high performance of *PetLock* through theory and simulation. Section 4 introduces various experiments on *PetLock*. Finally, section 5 concludes and discusses future work.

## 2. *PetLock* Design

*PetLock* is a multi-functional interface and is committed to becoming the space USB. *PetLock* has excellent capabilities, so that it can play a huge role in the OOC missions. We have summarized the main characteristics of *PetLock* (Table 2). *PetLock* mainly includes the following three interface functions:

- Mechanical Interface: Providing the functions of Misalignment tolerances, connection and mechanical loads transfer;

- Power Interface: Providing the capability of power transfer between modules;
- Data Interface: Providing the capability of data transfer between modules.

Table 2 The main characteristics of *PetLock*

| | |
|---|---|
| **Dimensions (diameter x height)** | Single：80 x 38mm<br>Combined：80 x 44mm（no rotor） |
| **Mass** | Single：0.3kg |
| **General features** | genderless，120° symmetry |
| **Misalignment tolerances** | +/- 12mm in translation<br>+/- 41° in rotation<br>+/- 14° in deflection |
| **Load transfer (nondestructive)** | 3000N in traction, 500Nm in rotation/bending |
| **Coupling sequence** | 10-20S（Adjustable） |
| **Power transfer** | 500W@48V<br>50W@24V |
| **Data transfer** | CAN, Ethernet |

### 2.1 Mechanical Interface and Locking Mechanism

Due to the specific 3D connection face design (Fig.4), *PetLock* has a large misalignment tolerance. The 3D face design comprehensively considers the misalignment tolerances of translation, rotation and deflection directions, which makes the *PetLock* do not need accurate visual recognition when docking. At the same time, the design of the special 3D face enables *PetLock* to transmit greater bending and rotation torque.

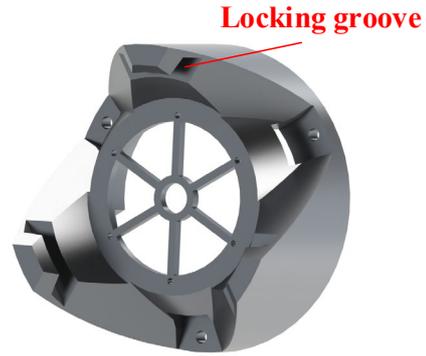

Fig.4 The specific 3D connection face

As Fig.5 shows, the locking mechanism features the three locking pins retraction structure design. The locking pins can provide a large locking force which improve the reliability of the interface. The three-point locking balances the contradiction between the reliability of locking and the difficulty of machining that's why adopt the design of three locking pins 120 ° evenly distributed. *PetLock* adopts the genderless design scheme, which not only increases the redundancy of the system hardware, but also can transmit greater mechanical loads when both sides of the interface are locked at the same time.

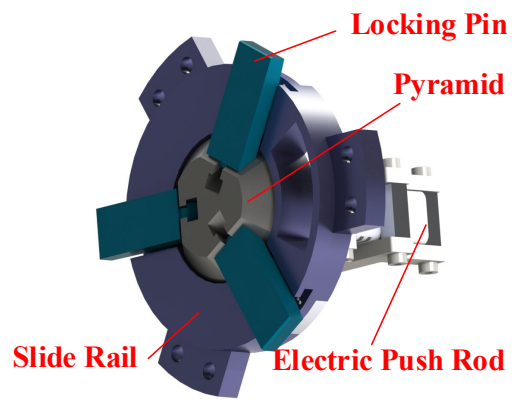

Fig.5 The Locking Mechanism

As Fig.5 shows, driven by the electric push rod, *Pyraimd* moves axially, pushes the locking pins to move radially,

and fits with the locking groove on the 3D face of the other side of interface, thus completing the locking of the interface. On the contrary, under the pulling action of the electric push rod, the interface is unlocked. The locking mechanism and the electric push rod are both self-locking, which can transfer the mechanical loads to the 3D face and the slide rail, thus ensuring the reliability of the whole mechanism.

### 2.2 Power and Data interface

*PetLock* transfers power and data through POGO pins. As shown in the Fig.6, POGO pins are uniformly arranged at 120 ° on the PCB. *PetLock* has two power interfaces, one for power transfer between robot joints and space facility modules, and the other for power transfer of *PetLock* electric push rod. It can transmit 500W @ 48V and 50W@ 24V respectively. *PetLock* also has two data interfaces, one is Ethernet, which is used to transmit data between robot joints and space facility modules, and the other is CAN, which is used to transmit data between *PetLocks*.

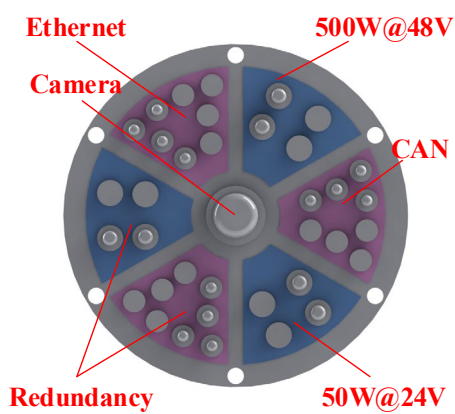

Fig.6 The power and data interface

## 3. Feasibility Analysis

In this part, the dynamic analysis of the locking mechanism is first carried out, and the self-locking property of the locking mechanism is proved, and then verified by simulation. The capability of *PetLock* 's mechanical loads transfer is also verified by simulation.

### 3.1 Dynamics Analysis

As Fig.7 shows that the force diagrams of the locking and unlocking processes of the locking mechanism, and the force analysis of the locking and unlocking processes is the same. The friction force on the locking pin can be expressed as,

$$f_1 = \mu_1 F_1$$
$$f_2 = \mu_2 F_2$$

Since the locking pins move to the right (left), the resultant force in the vertical direction is 0,

$$F_2 = f_1 \sin\theta + F_1 \cos\theta$$

The conditions that the locking pins' move ability can be expressed as,

$$F_1 \sin\theta > f_1 \cos\theta + f_2$$

After equivalent replacement,

$$\sin\theta > \mu_1 \cos\theta + \mu_1\mu_2 \sin\theta + \mu_2 \cos\theta$$

When $\mu_1 = \mu_2 = 0.3, \theta = 45°$,

$$1 > \mu_1 + \mu_1\mu_2 + \mu_2 = 0.69$$

So the locking mechanism can work normally. The dynamic simulation is carried out by Adams, and the simulation results conform to the theoretical analysis. As Fig.8 shows that the force curve of the locking pins during the locking and unlocking process.

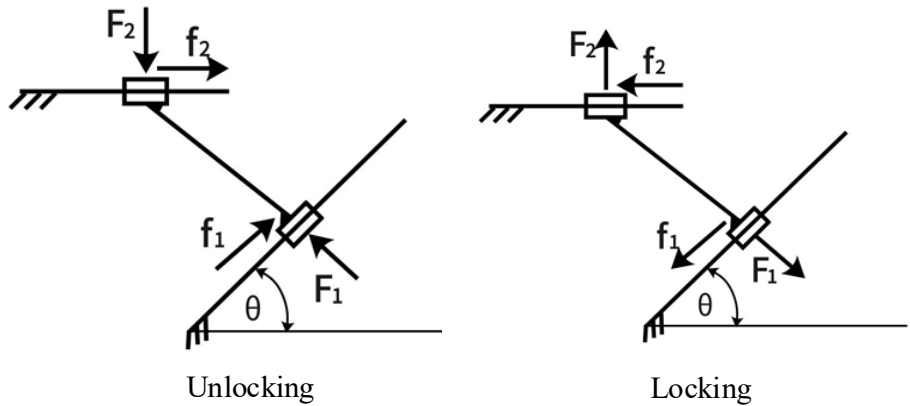

Unlocking　　　　　　　Locking

Fig.7 The force analysis diagrams

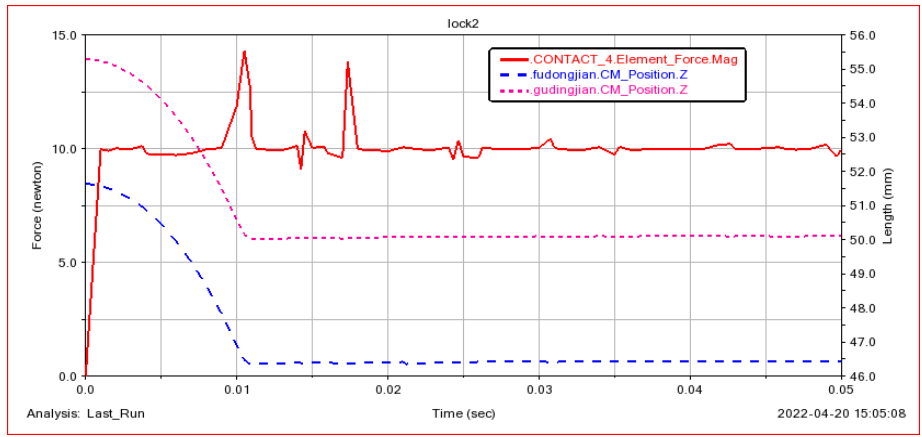

Locking process

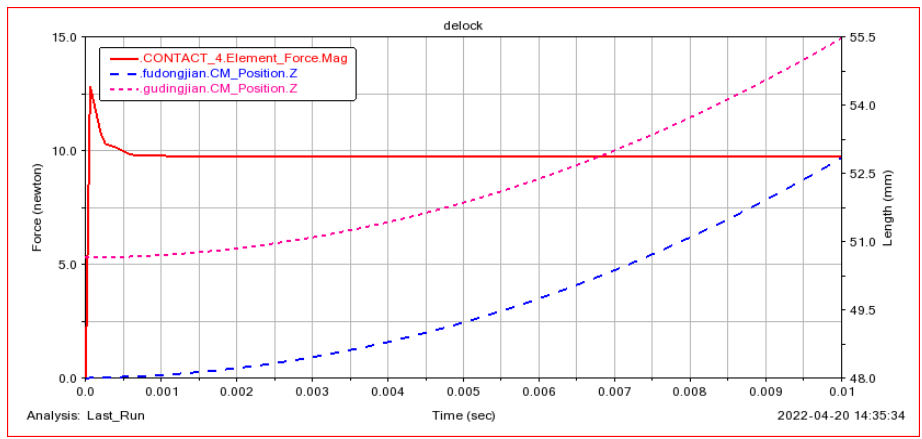

lnLocking process

Fig. 8 The force on locking pins

### 3.2 Self-Locking

The locking mechanism has the capability of self-locking, as Fig.10 shows that the condition of self-locking is,

$$f \cos \beta > F \sin \beta$$

And,

$$f = \mu F$$

So when $\mu > \tan \beta$, the locking mechanism has the capability of self-locking. When $\beta = 5°$, $\mu > \tan 5° = 0.0875$, the locking mechanism can lock self.

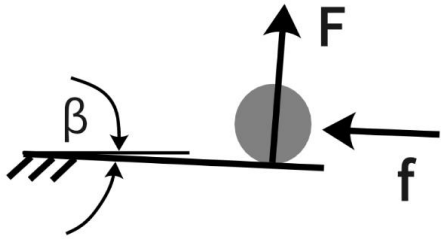

Fig.9 The force analysis diagrams when locking

### 3.3 Static Analysis

This section simulates the structural strength of ***PetLock*** when it is locked through the Static Analysis module of ANSYS. The 3D face and slide rail are made of 7075 aluminum alloy, and other parts are made of titanium alloy. The mechanical loads are 3000N tensile force, 500Nm rotational torque, 500Nm bending torque respectively and their combinations. The simulation results are summarized in Table 3. The results show that under the extreme conditions of simultaneously withstanding 3000N tensile force, 500Nm rotational torque and 500Nm bending torque, the maximum strain and stress of ***PetLock*** are within the allowable range, and the structural strength is reliable.

Table 3 ANSYS simulation results

| Condition | Maximum strain/mm | Maximum stress/MPa |
|---|---|---|
| **Tensile force 3000N** | 0.0037 | 21.999 |
| **Rotational torque 500Nm** | 0.0034 | 44.781 |
| **Bending torque 500Nm** | 0.0033 | 52.237 |
| **Tensile force 3000N Rotational torque 500Nm Bending torque 500Nm** | 0.0057 | 39.519 |

## 4. Experiment

This part introduces various experiments on ***PetLock***, including experiments on locking/unlocking process, power and data transfer, and misalignment tolerance experiments of 3D face.

### 4.1 Interface Function Verification

We first verified that the locking mechanism can lock/unlock normally. After that, we carried out experiments on the power and data transfer capability of the interface. The experimental results show that it can transfer 500W@48V and 50W@24V power at most, and it can realize CAN and Ethernet data transfer at the same time.

### 4.2 Misalignment Tolerance Experiments

As shown in the Fig.11, the misalignment tolerance experiments of 3D face are carried out using the UR3 robotic manipulator. One 3D face is fixed at the end of the robotic manipulator, and the other 3D face is placed on the desktop through a movable base with required degrees of freedom. When testing the translation and rotation misalignment tolerance, the 3D face on the desktop has two-DOF of translation and one-DOF of rotation. When testing the bending misalignment tolerance, the 3D face on the desktop has two-DOF of translation, one-DOF of rotation and two-DOF of deflection. Finally, the misalignment tolerance experiments in three directions was completed. The translation misalignment tolerance is 12mm, the rotation misalignment tolerance is 41 °, and the maximum deflection misalignment tolerance is 14 °.

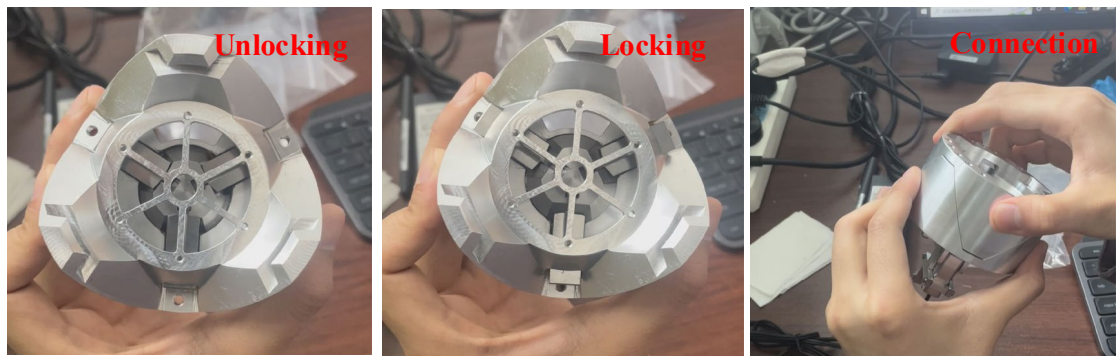

Fig.10 The Function Verification of *PetLock*

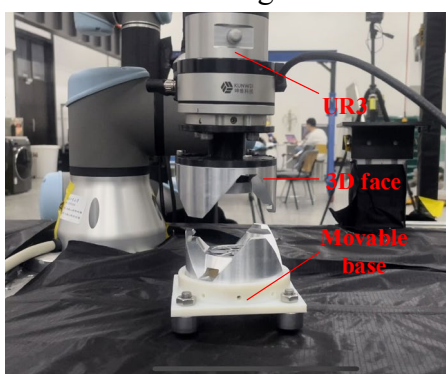

Fig.11 The misalignment tolerance experiments

## 5. Conclusion and Future Work

This paper presented the design and experiment of *PetLock*, a kind of genderless interface, which can transmit mechanical loads, power and data. The genderless design scheme increases its hardware redundancy. At the same time, the design of the specific 3D docking face and the three locking pins retraction locking mechanism makes it have the characteristics of large misalignment tolerance, high strength and high reliability on the premise of light weight. These excellent features will make *PetLock* shine in the future on orbit construction missions.

Our next work is to develop a reconfigurable robotic manipulator assembled and design some space facility modules using *PetLock*. And conducting ground test research on in orbit assembly.